# Uncertainty-Aware Large Language Models for Explainable Disease Diagnosis

Shuang Zhou[1], Jiashuo Wang[2], Zidu Xu[3], Song Wang[4], David Brauer[5], Lindsay Welton[5,6], Jacob Cogan[7], Yuen-Hei Chung[8], Lei Tian[9], Zaifu Zhan[10], Yu Hou[1], Mingquan Lin[1], Genevieve B. Melton[6,11], Rui Zhang[1,6,*]

**Affiliations:**
[1]Division of Computational Health Sciences, Department of Surgery, University of Minnesota, Minneapolis, MN, USA
[2]Department of Computer Science, University of Chicago, Chicago, IL, USA
[3]School of Nursing, Columbia University, New York, New York, USA
[4]Department of Electrical and Computer Engineering, University of Virginia, Charlottesville, VA, USA
[5]Division of Surgical Oncology, Department of Surgery, University of Minnesota, Minneapolis, MN, USA
[6]Center for Learning Health System Sciences, University of Minnesota, Minneapolis, MN, USA
[7]Department of Medicine, Division of Hematology, Oncology and Transplantation, University of Minnesota, Minneapolis, MN, USA
[8]Division of Cardiac Electrophysiology, University of California San Francisco, San Francisco, CA, USA
[9]Department of Biochemistry and Molecular Biology, Mayo Clinic, Rochester, MN, USA
[10]Department of Electrical and Computer Engineering, University of Minnesota, Minneapolis, MN, USA
[11]Institute for Health Informatics and Division of Colon and Rectal Surgery, Department of Surgery, University of Minnesota, Minneapolis, MN, USA

*Correspondence: ruizhang@umn.edu

## Abstract

Explainable disease diagnosis, which leverages patient information (e.g., signs and symptoms) and computational models to generate probable diagnoses and reasonings, offers clear clinical values. However, when clinical notes encompass insufficient evidence for a definite diagnosis, such as the absence of definitive symptoms, diagnostic uncertainty usually arises, increasing the risk of misdiagnosis and adverse outcomes. Although explicitly identifying and explaining diagnostic uncertainties is essential for trustworthy diagnostic systems, it remains under-explored. To fill this gap, we introduce ConfiDx, an uncertainty-aware large language model (LLM) created by fine-tuning open-source LLMs with diagnostic criteria. We formalized the task and assembled richly annotated datasets that capture varying degrees of diagnostic ambiguity. Evaluating ConfiDx on real-world datasets demonstrated that it excelled in identifying diagnostic uncertainties, achieving superior diagnostic performance, and generating trustworthy explanations for diagnoses and uncertainties. To our knowledge, this is the first study to jointly address diagnostic uncertainty recognition and explanation, substantially enhancing the reliability of automatic diagnostic systems.

# Introduction

Automatic disease diagnosis involves using computational models to predict the most likely diagnosis given observation data from patients (e.g., signs, symptoms, laboratory values). In real-world scenarios, merely providing diagnostic predictions often lacks trustworthiness due to the black-box nature of these models[1]. Some key features such as diagnostic explanations and adherence to clinical guidelines are essential to enhance trust and accuracy[2]. Specifically, diagnostic explanations allow clinicians to assess the model's reasoning[3], improving trust, while adherence to guidelines ensures standardization and reliability[4]. Incorporating these elements into diagnostic systems fosters greater trust and safety in clinical applications.

In practice, diagnostic uncertainty usually arises in clinical decision-making[5] particularly when clinical notes and other clinical data provide insufficient evidence, such as definitive symptoms or conclusive laboratory findings, posing challenges for reliable diagnoses[6]. For instance, a cross-sectional survey[6] of 32 primary care clinics reported that 13.6% of patients' consultations in primary care in the United States had missing clinical information. Another retrospective study[7] reported incomplete documentation of vital signs in nearly 48% of psychiatric emergency department cases. Identifying cases of diagnostic uncertainty enables clinicians to address gaps in evidence, minimizing misdiagnoses and adverse outcomes[6,8–10]. For example, an ICU patient presenting with chest pain, shortness of breath, and fatigue, but without definitive coronary angiography findings, may face diagnostic uncertainty between acute myocardial infarction[11] and Takotsubo cardiomyopathy[12]. Although these conditions share similar presentations, they differ markedly in etiology, severity, treatment, and outcomes[13]. In such cases, automatically identifying diagnostic uncertainty and explaining it can help prevent misdiagnosis, yielding considerable value.

Despite the development of advanced diagnostic systems[14,15], including those based on large language models (LLMs), current efforts struggle to recognize and explain diagnostic uncertainty[2,16]. Specifically, most systems rely on supervised deep learning trained on extensive labeled datasets[14,15,17]. While effective at distinguishing between diseases, these systems face challenges in incorporating learned medical knowledge, such as clinical guidelines, to assess the sufficiency of patient information or generate factual explanations[18,19]. Recently, LLM-based diagnostic systems have shown exceptional promise in clinical decision-making[16,20,21] by leveraging their generative capabilities and extensive medical knowledge to produce comprehensive diagnostic explanations[22]. These features position LLMs as valuable tools for recognizing and explaining diagnostic uncertainty. Pioneering studies[5,23] have begun exploring diagnostic uncertainty estimation. For example, Savage et al.[5] applied three commonly used confidence estimation methods[24]: confidence elicitation, token-level probability estimation, and self-consistency agreement to estimate the uncertainty degree of GPT[25] and LLaMA[26] models for disease diagnosis. However, these efforts face two critical limitations. First, LLM confidence often misaligns with factual accuracy[27], as LLMs may exhibit overconfidence in erroneous diagnoses due to flawed internal knowledge[28]. Second, these approaches[5,23] fail to provide narrative explanations for diagnostic uncertainty and the diagnoses, limiting their applicability in clinical practice.

Building an LLM-based diagnostic system that can effectively recognize and explain diagnostic uncertainty is challenging. In practice, clinicians are trained to adhere to diagnostic criteria[29], identifying uncertainties when criteria are unmet and seeking additional evidence to resolve them[30]. This adherence reflects a specialized human preference in medical decision-making[31]. While LLMs, including those tailored for medical use[32], are pre-trained on extensive biomedical corpora and fine-tuned for various tasks[33,34], they are generally not aligned with this professional preference[35,36]. Therefore, aligning LLMs to rigorously follow diagnostic criteria and recognize diagnostic uncertainty remains under-explored.

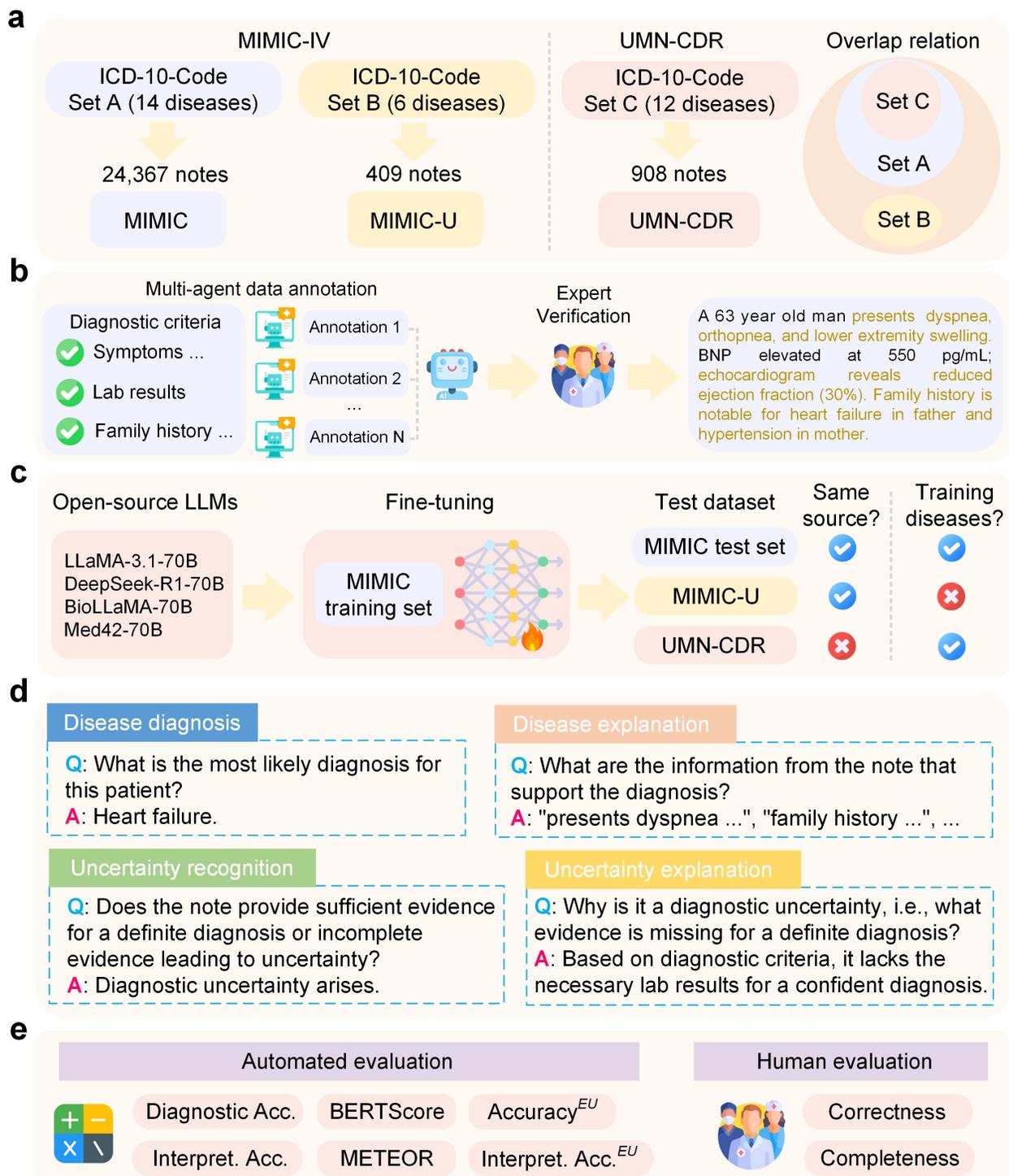

Fig. 1. Overview of dataset creation, annotation, and evaluation framework. **a.** To evaluate LLMs for disease diagnosis in realistic scenarios, we constructed three datasets using real-world clinical notes sourced from the MIMIC-IV and UMN-CDR databases. The disease types in the MIMIC dataset encompass those in the UMN-CDR dataset, whereas the MIMIC-U dataset features a distinct set of disease types. **b.** Diagnostic explanations supporting the diagnoses were annotated on clinical notes based on manually curated diagnostic criteria. A multi-agent framework was developed for annotation, with subsequent verification by medical experts. **c.** We fine-tuned open-source LLMs on the training data from the MIMIC dataset and assessed the performance on the MIMIC test set, MIMIC-U, and UMN-CDR datasets. Notably, MIMIC-U evaluates the model robustness on unseen diseases (i.e., absence from the training data), while UMN-CDR tests model generalizability across institutions. **d.** Given that the target problem involves diverse prediction types (see "Definition formulation" in the Methods), we divided the problem into four manageable subtasks for performance evaluation. **e.** Performance assessment involves various automated metrics complemented by human evaluation to ensure reliability.

This study aimed to address these gaps by developing uncertainty-aware LLMs for explainable disease diagnosis. Our contributions are threefold. First, we formally defined the task of uncertainty-aware diagnosis, enabling the recognition of uncertain cases alongside corresponding explanations. Second, we curated datasets with nuanced annotations to evaluate the trustworthiness of diagnostic models. Third, we proposed a tailored fine-tuning approach that integrated diagnostic criteria into the training process, aligning LLMs to rigorously adhere to these criteria for predictions. Building on open-source LLMs, we developed ConfiDx, which are customized diagnostic models to deliver accurate diagnoses while identifying and explaining diagnostic uncertainties. Experiments on the curated dataset demonstrated that ConfiDx outperformed off-the-shelf LLMs in diagnostic accuracy, enhanced recognition of diagnostic uncertainty, and provided reliable, comprehensive explanations. To the best of our knowledge, this is the first study to tackle this problem. Our work significantly advances the trustworthiness of LLM-based diagnostic models, securing reliable and explainable diagnoses.

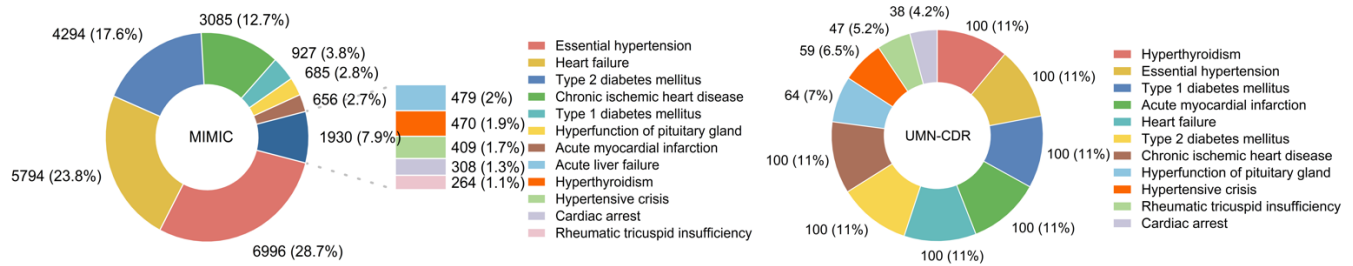

Fig. 2. Data composition of the constructed MIMIC and UMN-CDR datasets.

Table 1. Data statistics of the constructed MIMIC, MIMIC-U, and UMN-CDR datasets.

| Statistic | MIMIC | MIMIC-U | UMN-CDR |
|---|---|---|---|
| Total note number | 24,367 | 409 | 908 |
| Notes number in the training set | 17,057 | - | - |
| Notes number in the validation set | 2,436 | - | - |
| Notes number in the test set | 4,874 | 409 | 908 |
| Number of disease types | 12 | 6 | 11 |
| Total number of uncertainty cases | 12,183 | 204 | 454 |
| Mean note length (words) | 1,640 | 1,686 | 955 |
| Standard deviation of note length (words) | 489 | 482 | 343 |
| Mean number of explanations per note | 3.15 | 3.71 | 3.36 |
| Standard deviation of explanation number per note | 0.83 | 1.23 | 0.89 |

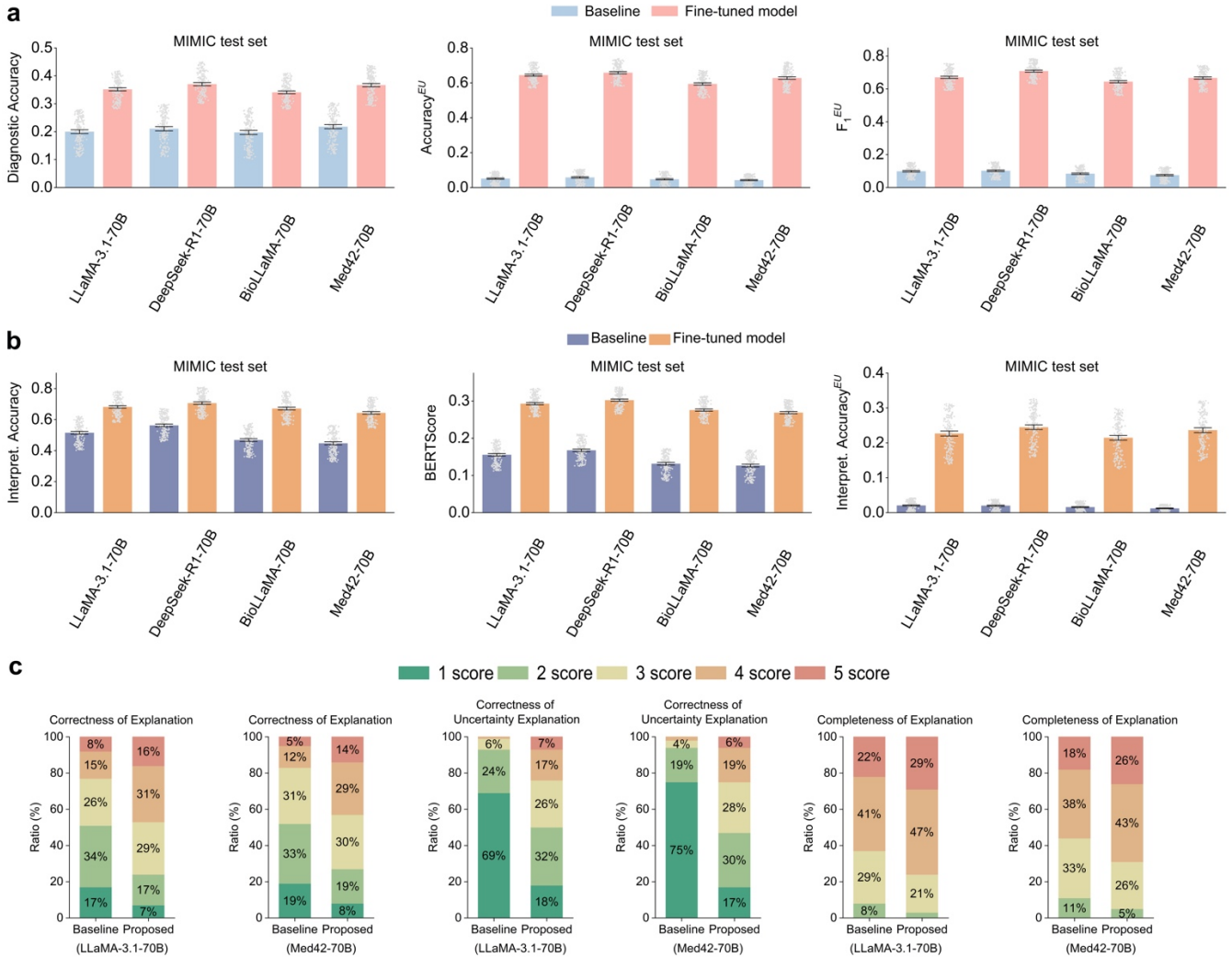

Fig. 3. Performance of disease diagnosis and diagnostic explanation on the test set of the MIMIC dataset. **a.** Performance comparison of baselines and the fine-tuned counterparts on disease diagnosis (measured by Diagnostic Accuracy) and diagnostic uncertainty recognition (measured by $Accuracy^{EU}$ and $F_1^{EU}$). Error bars represent the 95% CI of the mean, calculated via bootstrapping. **b.** Automatic evaluation of explanations for disease diagnosis (using metrics such as Interpret. Accuracy and BERTScore) and diagnostic uncertainty (using Interpret. $Accuracy^{EU}$). Error bars represent the 95% CI of the mean, calculated via bootstrapping. **c.** Results of manual evaluation of explanations, assessing the correctness and completeness of diagnostic explanations and the correctness of explanations for diagnostic uncertainty. Each metric was rated on a 5-point scale, where 1 represents the lowest and 5 is the highest score. Ratios below 3% are not highlighted in the figure.

# Results

## Dataset creation and evaluation framework

We constructed datasets using clinical notes from the MIMIC-IV[37] dataset and the University of Minnesota Clinical Data Repository (UMN-CDR) to develop and evaluate the proposed diagnostic models based on open-source LLMs (e.g., LLaMA). To assess model robustness on unseen disease types (i.e., absence from the training data), we created an independent test set, MIMIC-U, by holding out certain disease types from MIMIC-IV, leaving the rest as MIMIC. Additionally, we also collected publicly available case reports[38] from PubMed Central (PMC) to assess large-scale commercial LLMs, such as GPT-4o and Gemini[39], which generally cannot process privacy-sensitive clinical notes. We focused on three clinical specialties: endocrinology, cardiology, and hepatology, due to their significant impact on mortality in the U.S.[40] Dataset construction details are in Fig. 1 and the "Datasets" section of the Methods, with statistics in Table 1. Data composition for MIMIC, UMN-CDR, and MIMIC-U is shown in Fig. 2 and Supplementary Data 1.

In this study, we adopted four widely used open-source LLMs with 70 billion parameters (see "ConfiDx" section of the Methods). Given the complexity of uncertainty-aware disease diagnosis, which involves diverse prediction types (see "Definition formulation" in the Methods), we divided the problem into four manageable subtasks: disease diagnosis, diagnostic explanation, uncertainty recognition, and uncertainty explanation. We evaluated off-the-shelf and fine-tuned LLMs (ConfiDx), on these tasks, analyzing model robustness on hold-out disease types, generalizability across institutions, and key factors influencing performance.

## Uncertainty-aware diagnostic performance

The performance of disease diagnosis and diagnostic uncertainty recognition was evaluated on the MIMIC test set. The experimental prompts are detailed in Supplementary Data 2. As shown in Fig. 3a, the diagnostic accuracy of the off-the-shelf LLMs ranged from 0.197 (95% confidence interval (CI): 0.190–0.205) for BioLLaMA-70B to 0.218 (95% CI: 0.210–0.225) for Med42-70B. Fine-tuned models achieved accuracy improvements exceeding 68.3%, with all differences statistically significant ($p < 0.001$). For uncertainty recognition, the highest $Accuracy^{EU}$ and $F_1^{EU}$ scores among off-the-shelf LLMs were achieved by DeepSeek-R1-70B at 0.057 (95% CI: 0.054–0.062) and 0.102 (95% CI: 0.097–0.106), respectively. Remarkably, ConfiDx demonstrated substantial improvements, achieving $Accuracy^{EU}$ scores from 0.594 (95% CI: 0.587–0.600) for BioLLaMA-70B to 0.658 (95% CI: 0.652–0.665) for DeepSeek-R1-70B and achieving $F_1^{EU}$ scores from 0.644 (95% CI: 0.638–0.651) to 0.709 (95% CI: 0.702–0.715).

## Explanation performance

We evaluated the performance of the diagnosis and uncertainty explanation on the MIMIC test set, with prompts detailed in Supplementary Data 3. Partial results are presented in Fig. 2b, while additional metrics, including METEOR and SentenceBert, are provided in Supplementary Data 4. The interpretation accuracy of off-the-shelf LLMs ranged from 0.446 (95% CI: 0.437–0.456) for Med42-70B to 0.563 (95% CI: 0.554–0.571) for DeepSeek-R1-70B. Fine-tuning substantially improved performance, with gains ranging from 25.3% for DeepSeek-R1-70B to 43.8% for Med42-70B. For evaluation metrics such as BERTScore, SentenceBert, and METEOR, fine-tuned LLMs outperformed their off-the-shelf counterparts by an average of 96.7%, 63.4%, and 119.4%, respectively. Moreover, off-the-shelf LLMs demonstrated limited performance in uncertainty explanation, with an average performance of 0.017. In contrast, fine-tuned LLMs achieved superior accuracy, ranging from 0.214 (95% CI: 0.207–0.221) for BioLLaMA-70B to 0.245 (95% CI: 0.238–0.252) for DeepSeek-R1-70B, with performance improvements exceeding 0.2. All performance enhancements were statistically significant ($p < 0.001$).

Clinicians manually assessed explanations for diagnosis and uncertainty in terms of correctness[41] and completeness[42], as detailed in "Manual evaluation". Fig. 2c compares the performance of LLaMA-3.1-70B and Med42-70B. For diagnostic explanations, the correctness scores of off-the-shelf models predominantly ranged from 2 to 3, while fine-tuned models scored between 3 and 4. Regarding diagnostic uncertainty explanations, the two baselines received 69 and 75 scores of 1, while their fine-tuned counterparts achieved 50 and 53 scores above 2. In terms of completeness, the off-the-shelf Med42-70B achieved 38 scores of 4 and 18 scores of 5, while its fine-tuned counterpart achieved a superior explanation with 43 scores of 4 and 26 scores of 5.

## Robustness evaluation

We further evaluated model robustness using the holdout MIMIC-U dataset, which includes diseases that ConfiDx did not see from the training data. Fig. 4a presents the results for disease diagnosis and diagnostic uncertainty recognition. The diagnostic accuracy of ConfiDx ranged from 0.263 (95% CI: 0.257–0.270) for BioLLaMA-70B to 0.294 (95% CI: 0.287–0.300) for DeepSeek-R1-70B, representing performance improvements of 38.9%, 28.4%, 41.8%, and 39.3%, respectively, compared to their off-the-shelf counterparts. For diagnostic uncertainty recognition, fine-tuned models obtained an average $Accuracy^{EU}$ of 0.471 and an $F_1^{EU}$ of 0.497, marking substantial advancements over baseline performance of 0.046 on $Accuracy^{EU}$ and 0.083 on $F_1^{EU}$ ($p < 0.001$). The performance of the diagnostic explanation was also analyzed, as shown in Fig. 4b and Supplementary Data 5. Fine-tuned LLMs exhibited substantial improvements across all the explanation metrics, enhancing BERTScore by 60% and 49.2%, 56.2%, and 75.2%, respectively. Similarly, explanations for diagnostic uncertainty showed significant improvements, with an average score of 0.142. These ranged from 0.142 (95% CI: 0.135–0.149) for Med42-70B to 0.187 (95% CI: 0.180–0.194) for DeepSeek-R1-70B. These results demonstrate the ability of fine-tuned LLMs to perform effectively on the holdout MIMIC-U dataset.

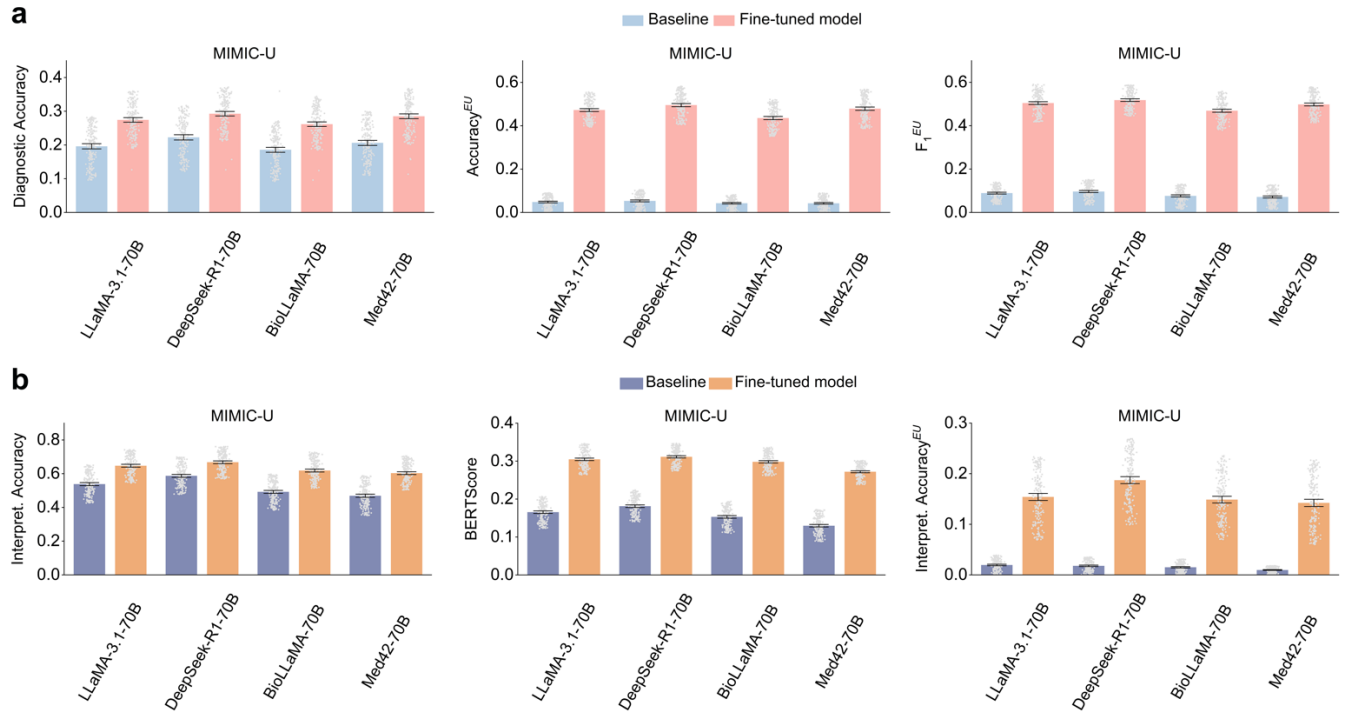

Fig. 4. Robustness evaluation performance on the MIMIC-U dataset with hold-out disease types. **a.** Performance on disease diagnosis (measured by Diagnostic Accuracy) and diagnostic uncertainty recognition (measured by $Accuracy^{EU}$ and $F_1^{EU}$). **b.** Automatic evaluation of explanations for diagnoses (using metrics such as Interpret. Accuracy and BERTScore) and for diagnostic uncertainty (using Interpret. $Accuracy^{EU}$). Error bars represent the 95% CI of the mean, calculated via bootstrapping.

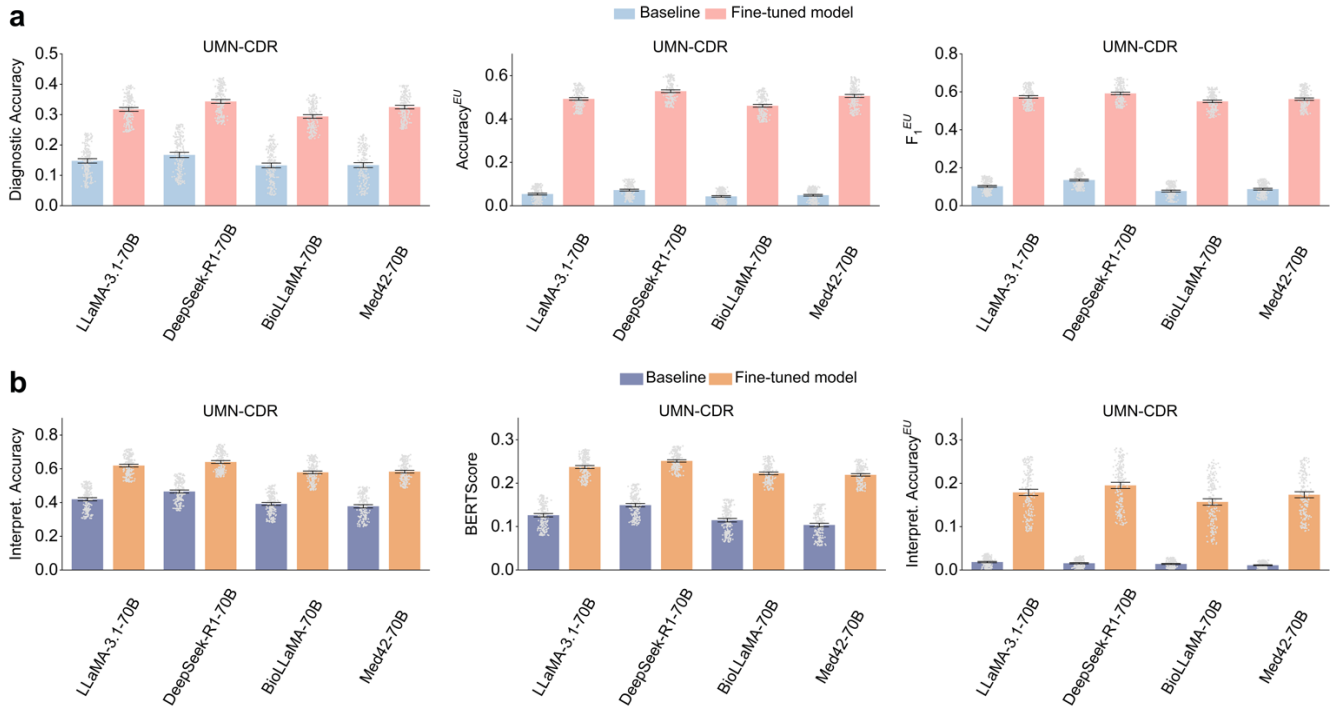

Fig. 5. Generalizability evaluation performance on the external UMN-CDR dataset. **a.** Performance on disease diagnosis (measured by Diagnostic Accuracy) and diagnostic uncertainty recognition (measured by $Accuracy^{EU}$ and $F_1^{EU}$). **b.** Automatic evaluation of explanations for diagnoses (using metrics such as Interpret. Accuracy and BERTScore) and for diagnostic uncertainty (using Interpret. $Accuracy^{EU}$). Error bars represent the 95% CI of the mean, calculated via bootstrapping.

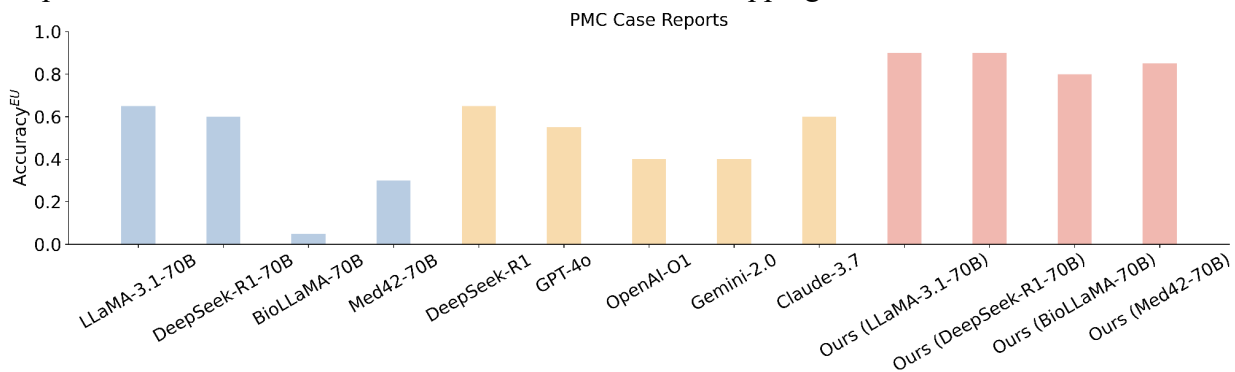

Fig. 6. Performance comparison with large-scale LLMs on diagnostic uncertainty recognition. The blue, yellow, and red color denotes baselines, large-scale LLMs (i.e., with hundreds of billion parameters), and our fine-tuned LLMs with 70 billion parameters, respectively. Our models with 70 billion parameters outperformed all the large-scale models.

## Generalizability evaluation

The model generalizability was evaluated through cross-site validation using the UMN-CDR dataset. Fig. 5a summarizes the results for disease diagnosis and diagnostic uncertainty recognition. Fine-tuned LLMs demonstrated significant improvements in diagnostic accuracy, with average increases of 0.169, 0.176, 0.162, and 0.186 across the four LLMs, respectively. For diagnostic uncertainty recognition, ConfiDx achieved an average $Accuracy^{EU}$ of 0.497 and $F_1^{EU}$ of 0.569, indicating substantial enhancements over their non-fine-tuned counterparts ($p < 0.001$). The explanation results are detailed in Fig. 5b and Supplementary Data 6, where ConfiDx consistently showed notable gains over the baselines across all explanation metrics ($p < 0.001$). For instance, fine-tuned LLaMA and DeepSeek improved BERTScore by 88.1% and 68.5%, respectively, and enhanced SentenceBert scores by 68.9% and 72.2%. Similarly, the performance of the diagnostic uncertainty explanation was significantly improved with an average of 0.162. The performance comparison shows that fine-tuned LLMs offer superior generalizability on the external dataset compared to their off-the-shelf counterparts.

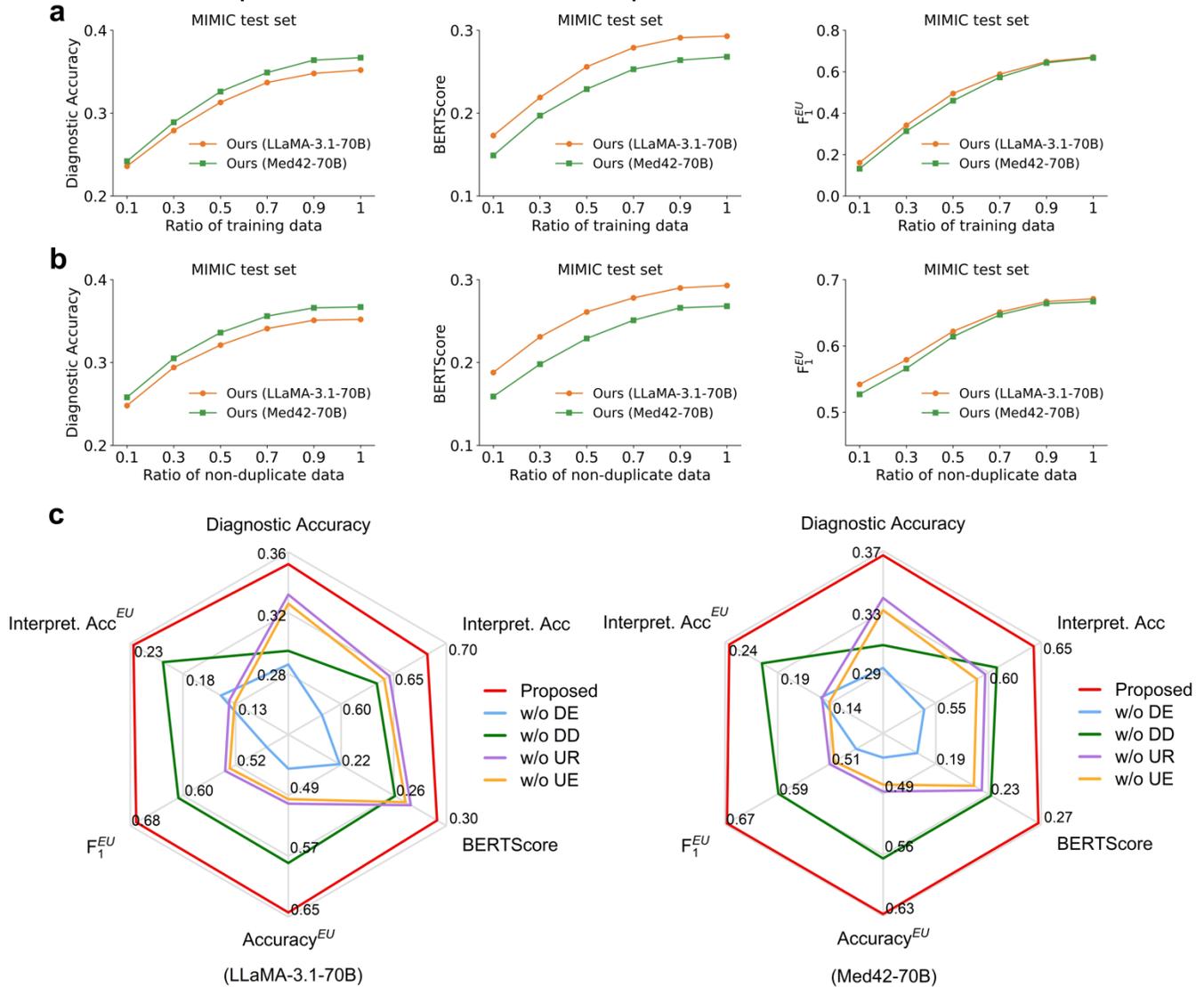

Fig. 7. The impact of several key factors on the overall performance. **a.** Performance with varying training data size, evaluated by randomly selecting portions of the training data ranging from 10% to 90%. **b.** Performance with varying data diversity in training data, achieved by randomly removing portions of the training samples while augmenting the remaining data to maintain a consistent training size. **c.** Ablation study performance with different learning objectives in the multi-task learning framework. For this, the learning objective of one task was removed from the current framework, and performance was evaluated across six metrics. Abbreviations: DE (diagnostic explanation), DD (disease diagnosis), UR (diagnostic uncertainty recognition), and UE (diagnostic uncertainty explanation).

## Comparison with large-scale LLMs

We evaluated the performance of large-scale commercial LLMs, comprising hundreds of billions of parameters, in recognizing diagnostic uncertainty. Their results were compared with ConfiDx on the PMC case reports. The analysis included five widely used LLMs, i.e., GPT-4o, OpenAI-O1, Gemini-2.0[39], Claude-3.7, and DeepSeek-R1[43]. As illustrated in Figure 6, the accuracy of large-scale LLMs ranged from 0.45 for OpenAI-O1 and Gemini-2.0 to 0.65 for DeepSeek-R1, with an average performance of 0.52, surpassing the average of 0.40 achieved by off-the-shelf LLMs with 70 billion parameters. However, ConfiDx with 70 billion parameters outperformed all large-scale models, delivering superior accuracy across the board, ranging from 0.80 to 0.90.

> **Ground-truth Diagnosis:** *Acute liver failure*
> **Ground-truth Explanation:** *The below evidence support the diagnosis {"Head CT showed diffuse cerebral edema", "INR-1.6", "developed the symptoms for less than 7 weeks"}*
> **Ground-truth Uncertainty Label:** *Insufficient information (Diagnostic uncertainty)*
> **Ground-truth Uncertainty Explanation:** *Lack of evidence on "No prior history of cirrhosis"*

> **(LLaMA-3.1-70B)**
> **Baseline - Predicted Diagnosis:** *Severe metabolic acidosis*
> **Baseline - Predicted Explanation:** *The below evidence support the diagnosis {"pH-7.20", "HCO3-12", "AnGap-27", "Base XS--13", "Lactate-11.4"}*
> **Baseline - Predicted Uncertainty Label:** *Sufficient information (Confident diagnosis)*
> **Baseline - Predicted Uncertainty Explanation:** *None*

> **(LLaMA-3.1-70B)**
> **Fine-tuned LLM - Predicted Diagnosis:** *Acute liver failure*
> **Fine-tuned LLM - Predicted Explanation:** *The below evidence support the diagnosis {"Head CT showed diffuse cerebral edema", "INR-1.6* BLOOD ALT-2705* AST-4966* TotBili-1.6", "developed the symptoms for less than 7 weeks"}*
> **Fine-tuned LLM - Predicted Uncertainty Label:** *Insufficient information (Diagnostic uncertainty)*
> **Fine-tuned LLM - Predicted Uncertainty Explanation:** *Insufficient evidence regarding "No prior history of cirrhosis"*

Fig. 8. Case study of off-the-shelf LLaMA-3.1-70B and the fine-tuned counterpart on the MIMIC dataset. Correct predictions are highlighted in blue.

## Impact of key factors on performance

We further investigated the influence of several key factors on the overall performance, including training data size, training data diversity, and the learning objective, on LLaMA-3.1-70B and Med42-70B. To assess the effect of training data size, we randomly selected portions of the training data, ranging from 10% to 90%, and evaluated the performance. As depicted in Fig. 7a, with merely 10% of the training data, the model achieved Diagnostic Accuracy, Interpretation Accuracy, and $Accuracy^{EU}$ scores of 0.239, 0.541, and 0.147, respectively. As the data volume increased, performance improved markedly before gradually converging. To investigate the impact of training data diversity, we randomly removed portions of the training samples while augmenting the remaining data to maintain a consistent training size. As shown in Fig. 7b, when only 10% of the original data was retained, ConfiDx achieved Diagnostic Accuracy, Interpretation Accuracy, and $Accuracy^{EU}$ scores of 0.278, 0.586, and 0.533, respectively. Performance steadily improved with increasing data diversity and similarly converged as the diversity increased. Additionally, we assessed the impact of the learning objective within the multi-task learning framework. Specifically, we removed one task at a time from the original learning objective and measured the performance across six metrics. The results in Fig. 7c demonstrated that each task within the framework contributes meaningfully to the model's overall performance.

## Case study

Case studies were conducted to demonstrate the superior explainability of the fine-tuned LLMs over their off-the-shelf counterpart. As shown in Fig. 8, off-the-shelf LLaMA produced an incorrect diagnosis and flawed explanations, whereas the fine-tuned model made the correct diagnosis along with three accurate explanations and managed to recognize and explain diagnostic uncertainty. Additional examples are provided in Supplementary Data 7.

# Discussion

In this study, we developed ConfiDx by fine-tuning open-source LLMs for uncertainty-aware disease diagnosis, addressing the dual challenges of conducting explainable diagnosis and recognizing diagnostic uncertainty due to insufficient evidence in clinical notes. Given the complexity of this task, which involves multiple prediction types, we divided it into four manageable subtasks for performance evaluation. We highlight the following key observations and insights for further discussion.

First, off-the-shelf LLMs demonstrated limited capability in diagnosing diseases, identifying diagnostic uncertainty, and explaining uncertainty across the three datasets (Figs. 3-5). The suboptimal performance is unsurprising, considering the challenges posed by real-world clinical notes[2], which often include extraneous information, such as past medical history, and overlapping disease presentations[3], such as cardiac arrest and atrial fibrillation. The unsatisfied performance in identifying diagnostic uncertainty stems from two primary factors: (1) Without pinpointing the most likely diagnosis, LLMs struggle to identify cases with incomplete evidence, as diagnostic criteria differ across diseases. (2) Off-the-shelf LLMs tend to overestimate the sufficiency of clinical note information[16,44], assuming it contains all necessary evidence for diagnosis, even when this contradicts explicit diagnostic criteria. This overconfidence likely results from their general-purpose training[25,26], which focuses on simpler tasks like text summarization and question answering. In contrast, real-world disease diagnosis is a complex endeavor requiring customized training to instill the medical expertise needed to follow diagnostic criteria rigorously and seek adequate supporting evidence[2,45]. This discrepancy underscores the necessity of adapting general-purpose LLMs for domain-specific applications to achieve expert-level performance[41]. Furthermore, the inability of off-the-shelf LLMs to explain diagnostic uncertainty is reasonable, as accurate diagnosis and uncertainty recognition are prerequisites for meaningful explanation.

Second, fine-tuned LLMs significantly outperformed their off-the-shelf counterparts in diagnosis and uncertainty recognition on the MIMIC test set (Fig. 3a). The enhanced diagnostic accuracy stems from extensive training data, which allowed the models to learn diverse disease patterns, such as characteristic signs, symptoms, and laboratory results, improving their ability to distinguish between similar conditions. Additionally, the fine-tuning process aligned the models with domain expert behavior by grounding their reasoning in diagnostic criteria. By simulating real-world scenarios where clinical notes lack complete evidence, fine-tuning enabled ConfiDx to assess the sufficiency of the information more effectively, thereby enhancing their ability to recognize uncertainty.

Third, ConfiDx excelled in explanatory capabilities for both diagnosis and uncertainty (Figs. 3b-3c) on the MIMIC test set, largely due to the injection of medical knowledge via instruction fine-tuning. By explicitly mapping diagnostic criteria to varied symptom descriptions, the annotated data enabled ConfiDx to understand the meaning of the diagnostic criteria thoroughly. For instance, the models learned to associate diagnostic criteria like "Related symptoms: polyuria, polydipsia, and unexplained weight loss" with variable patient descriptions, such as "three weeks of polyuria and polydipsia" or "10-month history of polyuria with a 15 kg decrease in body weight." This alignment with diagnostic criteria enabled ConfiDx to identify unmet criteria accurately and clarify the basis for uncertainty.

Additionally, we discovered that ConfiDx exhibited fair robustness in diagnosing diseases, recognizing diagnostic uncertainty, and providing explanations for diseases absent from the training data (Fig. 4). This suggests that the fine-tuning approach effectively leveraged the LLMs' internal knowledge, grounded in diagnostic criteria, to recognize uncertainty and deliver explanations for unseen cases. Such robustness is particularly significant to clinical applications[15,46], as it is generally impractical to annotate data for all possible disease types[17]. Encouraging LLMs to effectively leverage the extensive parametric knowledge offers a cost-effective solution in resource-constrained settings.

The fine-tuned LLMs also exhibited superior generalizability, outperforming off-the-shelf models on the external UMN-CDR dataset (Fig. 5). Because the MIMIC training data and the UMN-CDR dataset differ in several critical aspects, such as information recording styles and text length (as outlined in Table 1). This superior generalization underscores the ability of ConfiDx to internalize relevant medical knowledge from diagnostic criteria, enabling their adaptation to clinical notes from diverse healthcare systems.

Several factors influence performance in this study, including the size and diversity of the training data and the choice of learning objectives. Specifically, larger training data exposed the models to a broader spectrum of disease patterns, thus improving diagnostic accuracy[20] (Fig. 7a). Similarly, more data with detailed evidence-based explanations strengthened the models' ability to associate diagnostic criteria with varying symptom descriptions, thereby enhancing explanation quality. Besides, data diversity was equally critical (Fig. 7b), as ConfiDx depends on exposure to comprehensive patterns to handle real-world variability[47]. However, performance plateaued with sufficient data coverage (e.g., 90% of the training set, Figs. 7a-7b), suggesting diminishing returns from expansions in data size or diversity. Thus, carefully considering data size, diversity, and the cost-effectiveness trade-off is essential when constructing fine-tuning datasets. Additionally, as shown in Fig. 7c, each learning objective contributes to the overall performance in disease diagnosis, uncertainty recognition, and explanation, reflecting the subtasks' necessity.

Despite these advances, our study has the following limitations. First, we did not evaluate smaller LLMs, such as MedAlpaca-7B[48] or PMCLLaMA-13B[49] for two key reasons. These models typically have limited in-context lengths, rendering them inadequate for processing real-world clinical notes, which often contain extensive text. Additionally, large-scale models are generally preferred for real-world applications due to their superior performance[1,50]. Smaller models often lack sufficient clinical knowledge or robust instruction-following capabilities[2,51], limiting their applicability in evidence-based diagnosis. Second, while we observed that closed-source models, such as GPT-4o, also struggled with uncertainty recognition, we were unable to validate the proposed approach on these large-scale commercial models.

In summary, our study highlights the critical limitations of current LLMs in recognizing and explaining diagnostic uncertainty. To address this, we formalized the problem of uncertainty-aware disease diagnosis and constructed customized datasets with nuanced annotations for training and evaluation of the trustworthiness of diagnostic models. Besides, we integrated diagnostic criteria into the fine-tuning and developed ConfiDx capable of diagnoses, uncertainty recognition, and delivery explanations. Our fine-tuned LLMs outperformed the off-the-shelf counterparts, achieving superior diagnostic accuracy, more reliable explanations, robust recognition of uncertain cases, and fair generalization in cross-site evaluations. This study significantly advances the trustworthiness of LLM-based diagnostic models, alleviating concerns about diagnostic uncertainty raised in clinical practice, and securing reliable and explainable clinical decision-making.

## Methods

### Definition formulation

**Evidence-Based Diagnosis:** It refers to a diagnosis made by adhering to established diagnostic criteria, utilizing a patient's clinical information as the basis for decision-making.
**Confident Diagnosis:** It builds on evidence-based diagnosis, signifying a high degree of certainty in the diagnostic decision.
**Diagnostic Uncertainty:** It arises when diagnostic criteria are unmet due to insufficient evidence, such as the absence of definitive symptoms, ambiguous clinical signs, or inconclusive laboratory findings.
**Uncertainty-Aware Disease Diagnosis:** This task involves developing predictive models that leverage clinical information to determine both the most likely diagnosis and occurrence of diagnostic uncertainty. The models distinguish between confident diagnoses (supported by sufficient evidence) and those characterized by insufficient evidence while providing detailed explanations.

### Datasets

The MIMIC-IV database[37] provides de-identified electronic health records (EHRs) for nearly 300,000 patients treated at Beth Israel Deaconess Medical Center in Boston, Massachusetts, USA, from 2008 to 2019. This publicly accessible dataset includes diverse clinical information such as laboratory results, diagnoses, procedures, and unstructured notes like discharge summaries and radiology reports. The UMN-CDR database, a private resource, contains EHRs for approximately 400,000 patients treated at the University of Minnesota Medical Center, USA, from 2008 to 2022. These records include laboratory results, treatments, and diagnoses, offering extensive clinical information. PubMed Central, maintained by the U.S. National Center for Biotechnology Information, is a free digital archive of biomedical literature. It provides millions of full-text research articles spanning topics like clinical studies, case reports, and systematic reviews, ensuring comprehensive access to scientific publications.

We focused on three clinical specialties linked to critical mortality factors in the USA[40]: endocrinology, cardiology, and hepatology. Diseases were selected through the following process: (1) identifying all diseases within each specialty based on ICD-10-CM codes; (2) excluding diseases without clear diagnostic criteria; (3) removing diseases with fewer than two rules; and (4) excluding diseases not appearing as the primary diagnosis in the selected notes. This yielded 18 disease types, as shown in Supplementary Data 1. The note selection and pre-processing are detailed in Supplementary Data 12. To assess the robustness of LLMs for unseen diseases, we created an independent test set, MIMIC-U, by holding out notes for six diseases with limited notes. The remaining dataset was referred to as MIMIC. For evaluating model generalizability, we extracted notes from the UMN-CDR dataset corresponding to the same disease types as in the MIMIC dataset. This dataset included 11 disease types, excluding acute liver failure. If a disease type has over 100 notes, we randomly sampled 100 notes. From the PMC dataset, we selected 20 real-world case reports tagged as "case report". Data statistics of the MIMIC, MIMIC-U, and UMN-CDR are in Table 1.

## Data annotation and split

**Diagnostic Criteria.** Diagnostic criteria established by professional associations serve as the gold standard for clinical diagnoses. We manually collected these criteria from authoritative guidelines (detailed in Supplementary Data 8) such as the American Diabetes Association's criteria for Type 1 Diabetes Mellitus[52] and the American Heart Association's criteria for Acute Myocardial Infarction[53].

**Annotation Process.** Clinical notes were annotated to evaluate the ability of LLMs to identify diagnostic uncertain cases. Diagnostic criteria served as proxies to determine whether sufficient evidence was present for a confident diagnosis. Specifically, notes containing sufficient evidence aligned with the criteria were labeled as evidence-complete, while those with insufficient evidence were marked as uncertain.

To balance annotation accuracy and efficiency, we employed a multi-agent framework (see Supplementary Data 9) for the majority of the annotation on MIMIC-IV and UMN-CDR datasets, following established methodologies[3,54]. This framework effectively annotated the evidence within the notes, grounded in diagnostic criteria. All relevant descriptions were annotated as diagnostic evidence to produce evidence-complete notes. Evidence-incomplete notes were generated by masking portions of the evidence, simulating incomplete clinical information. Domain experts reviewed the modified notes to ensure that masking did not alter the diagnosis. We also sampled a subset of notes for human annotation to evaluate inter-annotator agreement (IAA) between the LLM and human annotations (Supplementary Data 9). For the PMC dataset, two experts reviewed each report, extracted diagnoses, and annotated all evidence supporting the diagnosis based on diagnostic criteria.

**Data Split.** Figure 1 illustrates the data split, with notes maintaining a 1:1 ratio of evidence-complete to incomplete cases. The dataset was randomly divided into training, validation, and test sets in a 7:1:2 proportion while preserving this balance across all subsets.

## Comparison with other large-scale LLMs

We aim to verify whether advanced commercial LLMs with hundreds of billions of parameters perform well in uncertainty-aware diagnosis and compare them with our fine-tuned 70B-parameter LLMs. Due to privacy concerns, we assessed model performance on publicly available case reports from the PMC portal. We tested five state-of-the-art LLMs, GPT-4o (gpt-4o-2024-08-06), OpenAI-O1 (o1-2024-12-17), Gemini-2.0 (gemini-2.0-flash), and Claude-3.7 (claude-3-7-sonnet-20250219), and DeepSeek-R1, focusing on diagnostic uncertainty recognition. For a fair comparison, each model was provided with the ground-truth diagnosis and corresponding diagnostic criteria, mirroring the data annotation process. Using zero-shot prompting (Supplementary Data 10), we instructed the LLMs to determine whether the clinical notes contained sufficient evidence for a confident diagnosis. A clinician manually reviewed all predictions.

## ConfiDx

We developed ConfiDx on open-source LLMs for explainable diagnosis via instruction fine-tuning, where diagnostic tasks were formatted as natural language instructions with structured outputs. Clinical notes were converted into instructional demonstrations, each comprising: (1) an input note with patient information, (2) a task instruction, and (3) an output containing the diagnosis, explanation, and uncertainty label.

A key challenge arose from the need to process complex, multi-component instructions for simultaneous prediction of diagnosis, explanation, and uncertainty. To address this, we decomposed the task into four subtasks: (1) disease diagnosis, (2) diagnostic explanation, (3) uncertainty recognition, and (4) uncertainty explanation. This decomposition enabled more effective processing through a multi-task learning framework, with separate instruction sets for each component (Supplementary Data 11). For optimization, we used cross-entropy loss for diagnosis and uncertainty recognition. For the explanation generation, we employed soft F1 loss[55,56] to account for potential discrepancies between the number of explanatory components in predictions and references, and the need for an item-by-item comparison.

We adopted four widely used open-source LLMs for training and evaluation on real-world clinical notes: two general-domain models, LLaMA-3.1-70B-Instruct (LLaMA-3.1-70B) and DeepSeek-R1-Distill-Llama-70B[43] (DeepSeek-R1-70B), and two domain-specific models, BioLLaMA-70B and Med42-70B[57]. We selected models with 70 billion parameters to balance computational efficiency and predictive performance, as clinicians often favor larger models for their reliability and trustworthy predictions in complex diagnostic scenarios[1,50,58]. We implemented parameter-efficient fine-tuning via Low-Rank Adaptation (LoRA)[59] using the Hugging Face framework. Experiments were conducted with a batch size of 4, 10 epochs, a maximum learning rate of 1e-4, and a 0.01 warm-up ratio. LoRA parameters included a rank of 8, alpha of 16, and a dropout rate of 0.05. No further hyperparameter tuning was performed. Fine-tuning and inference were performed on 10 Nvidia A100 GPUs (80 GB VRAM) with input data and instructions provided during inference.

## Key factors analysis

To evaluate the impact of training data size, we varied the training set from 10% to 90% of the data for fine-tuning, with notes randomly selected while stratifying by disease type. For data diversity analysis, we reduced diversity by retaining only 10% to 90% of the data, stratified by disease type, and duplicating the remaining notes to restore the original dataset size. This process ensured the data size remained consistent despite reduced diversity. To analyze the effect of learning objectives, we excluded one task at a time from the original objective, fine-tuned the model with the modified objective, and evaluated performance.

## Automatic evaluation metrics

Following related works[1,2,60,61], we evaluated diagnostic performance with accuracy and the ability to correctly identify uncertainty with $Accuracy^{EU}$ and $F_1^{EU}$ score. The computation of $Accuracy^{EU}$ is as follows:

$$Accuracy^{EU} = \frac{\text{Cumulative number of correctly recognized evidence uncertainty}}{\text{Total number of evidence uncertainty}} \quad (1)$$

The computation of $F_1^{EU}$ score is shown below:

$$\text{Precision} = \frac{\text{TP}}{\text{TP} + \text{FP}} \quad (2)$$

$$\text{Recall} = \frac{\text{TP}}{\text{TP} + \text{FN}} \quad (3)$$

$$F_1^{EU} = \frac{2 \times \text{Precision} \times \text{Recall}}{\text{Precision} + \text{Recall}} \quad (4)$$

where TP (true positives) denotes uncertainty cases correctly identified as uncertain, FP (false positives) denotes cases incorrectly identified as uncertain, and FN (false negatives) refers to missed or unrecognized uncertainty cases. Regarding explanation evaluation, we adopted METEOR[62], BERTScore[63], SentenceBert[64], and Interpretation Accuracy[3] since these metrics have been widely used to measure the semantic similarity between the reference and generated text[65,66]. Concretely, METEOR[62] primarily focused on surface-level similarities by leveraging n-gram overlap and linguistic features like stemming and synonyms, while BERTScore[63] and SentenceBert[64] assessed semantic similarity using contextual embeddings from pre-trained language models. The Interpretation Accuracy[3] is computed as:

$$\text{Interpret. Accuracy} = \frac{\text{Cumulative number of correct explanations}}{\text{Total number of explanations}} \quad (5)$$

Similarly, we measured the alignment between the generated explanations and ground-truth with $Interpret.Accuracy^{EU}$ using the formula:

$$Interpret.Accuracy^{EU} = \frac{\text{Cumulative number of correct explanations for incomplete evidence}}{\text{Total number of explanations for incomplete evidence}} \quad (6)$$

Mean values and 95% confidence intervals for all metrics were estimated through a non-parametric bootstrap procedure with 200 iterations. For each bootstrap iteration, a resampled dataset of the same size as the test set was generated through random sampling with replacement.

## Manual evaluation

We conducted a manual evaluation of the predicted explanations. For diagnostic explanations, we randomly selected 100 notes from the MIMIC test set, evenly split between evidence-complete and evidence-incomplete notes. For uncertainty recognition explanations, we randomly sampled 100 evidence-incomplete notes from the same source. Following prior work[41,42], we assessed explanation quality along two dimensions: correctness, reflecting medical accuracy, and completeness, measuring how comprehensively symptom descriptions were addressed. Each metric was rated on a 5-point scale. A score of 5 indicated that over 80% of ground-truth explanations were accurately predicted (correctness) or that more than 80% of explanations matched the ground-truth (completeness). A score of 1 denoted less than 20% accuracy or completeness. Two physicians independently graded each note, with a third resolving any discrepancies.

# Author Contributions

S.Z. and R.Z. conceptualized the study. S.Z. led the design of the study. S.Z., J.W., and Z.X. conducted the literature search, while S.Z., J.W., and S.W. were responsible for model construction. S.Z. and R.Z. discussed and organized data annotation and human evaluation. D.B., L.W., J.C., Y.C., L.T., S.Z., Z.X., and Z.Z. contributed data collection, preprocessing, annotation, or statistics. S.Z., M.L., Y.H., and R.Z. developed the experimental design. S.Z. drafted the initial manuscript, and R.Z. provided supervision throughout the study. All authors participated in research discussions, critically revised the manuscript, and approved the final version for submission.

# Acknowledgment

This work was supported by the National Institutes of Health's National Center for Complementary and Integrative Health under grant number R01AT009457, National Institute on Aging under grant number R01AG078154, and National Cancer Institute under grant number R01CA287413. The content is solely the responsibility of the authors and does not represent the official views of the National Institutes of Health. We also acknowledge the support from the Center for Learning Health System Sciences.

## Competing Interests

The authors declare no competing interests.